\documentclass{article}

\usepackage{PRIMEarxiv}

\usepackage[utf8]{inputenc} 
\usepackage[T1]{fontenc}    
\usepackage{hyperref}       
\usepackage{url}            
\usepackage{booktabs}       
\usepackage{amsfonts}       
\usepackage{nicefrac}       
\usepackage{microtype}      
\usepackage{fancyhdr}       
\usepackage{graphicx}       
\graphicspath{{./}}     

\usepackage{caption}
\usepackage{array}
\usepackage{adjustbox}
\hypersetup{
colorlinks=true,
urlcolor=blue
}

\title{A conditional Generative Adversarial network model for the Weather4Cast 2024 Challenge}

\author{
  Atharva Deshpande, Kaushik Gopalan, Jeet Shah, and Hrishikesh Simu \\
  The Centre for inter-disciplinary Artificial Intelligence \\
  FLAME University \\
  Pune\\
  \texttt{\{atharva.a.deshpande, kaushik.gopalan, jeet.c.shah, hrishikesh.simu\}@flame.edu.in} \\  
}

\begin{document}
\maketitle

\vspace{-8pt}

\begin{abstract}
This study explores the application of deep learning for rainfall prediction, leveraging the Spinning Enhanced Visible and Infrared Imager (SEVIRI) High rate information transmission (HRIT) data as input and the Operational Program on the Exchange of weather RAdar information (OPERA) ground-radar reflectivity data as ground truth. We use the mean of 4 InfraRed frequency channels as the input. The radiance images are forecasted up to 4 hours into the future using a dense optical flow algorithm. A conditional generative adversarial network (GAN) model is employed to transform the predicted radiance images into rainfall images which are aggregated over the 4 hour forecast period to generate cumulative rainfall values. This model scored a value of approximately 7.5 as the Continuous Ranked Probability Score (CRPS) in the Weather4Cast 2024 competition and placed 1\textsuperscript{st} on the core challenge leaderboard. 
\end{abstract}


\section{Introduction}
Accurate weather prediction, particularly rainfall forecasting, is essential for agriculture, transportation, disaster management, and urban planning. However, forecasting rainfall is a complex task due to its dynamic and unpredictable nature. Recent advancements in data-driven methods, including deep learning, offer new ways to tackle this challenge by utilizing large meteorological datasets to identify and predict spatial and temporal patterns in weather systems \cite{menzel_application_1998, zhang_improving_2019}.

This study is part of the Weather4Cast Challenge, a competition designed to improve rainfall forecasting using high-resolution weather data \cite{gruca_weather4cast_nodate}. The challenge focuses on forecasting satellite-based visible and infrared images into high-resolution precipitation rate maps. To address this, we use a model for translating images to forecast rainfall several hours ahead.

Optical flow based models are currently operational as part of nowcast warning systems across the world \cite{shukla_satellite-based_2017, woo_operational_2017, sideris_nowprecip:_2020}. There are multiple methods and algorithms to perform optical flow-based nowcasting \cite{li_subpixel-based_2018, sideris_nowprecip:_2020, zhu_rain-type_2022}.These methods excel in short-term predictions by analyzing temporal and spatial motion patterns of weather systems. Notably, optical flow has been used in tandem with machine learning models to improve the accuracy of nowcasting systems. Beyond optical flow, deep learning techniques have shown significant promise in nowcasting applications. Models such as convolutional neural networks (CNNs) and recurrent neural networks (RNNs), including Long Short-Term Memory (LSTM) networks, have been employed to predict precipitation and other meteorological variables from satellite imagery and radar data \cite{shi_convolutional_2015, ravuri_skillful_2021}. These architectures are well-suited to capturing spatial and temporal dependencies in weather patterns, producing high-resolution, physically consistent forecasts. These two types of nowcasting techniques have also been made to work in tandem for many nowcasting models \cite{ha_deep_2023, nie_ofaf-convlstm:_2021}. 

In the following section, we outline our methods, including data preparation, model design, and the training process.

\section{Methodology}
\label{sec:methodology}
The Weather4Cast competition provides data from the SEVIRI images as input and rainfall estimates from OPERA as the target \cite{schmetz_geostationary_2002,saltikoff2019opera,huuskonen2014operational}. The OPERA estimates, which have 6 times the spatial resolution of the SEVIRI, are resampled to the SEVIRI resolution to create matching images of $252 \times 252$ pixel size. The problem being posed at the competition is to use SEVIRI images for 1 hour -- i.e. 4 consecutive SEVIRI images - to predict the cumulative rain rate aggregated over the subsequent 4 hours. We approach this problem in 2 parts: 1) we first use dense optical flow to forecast SEVIRI images for the subsequent 4 hours and 2) we then apply a cGAN to transform the forecasted radiance images into corresponding rainfall values at the SEVIRI resolution. These coarse rainfall estimates are then resampled to OPERA resolution using bilinear interpolation.

While there are 11 channels from SEVIRI available for use, not all of them are suitable for use in the estimation of rainfall. The primary indicator of the likelihood of rainfall at a given pixel is the presence of cold cloud-top temperatures at the location. We reject the Near-InfraRed and Water Vapor channels because they are not strongly related to the cloud top temperatures. The Visible frequency channels are strong indicators of precipitation, but they are only available in the daytime. While it would be preferable to use flexible models that can use the Visible channels during daytime and use a more limited dataset during night, this was beyond the scope of our abilities in the limited time available for the competition. As a result, we discarded the Visible channels from use as inputs as well. Among the InfraRed frequency channels, the 2 window channels at 10.8 $\mu m$ and 12.0 $\mu m$ represent the most reliable information about cloud-top temperatures. In this study, we focus on these 2 channels allong with the 2 absorption channels neighboring them; i.e. the channels at 9.7 $\mu m$ and 13.4 $\mu m$. Table~\ref{tab:correlation_matrix} shows the correlation between each pair of these channels derived from 100 consecutive images in the training dataset; i.e. $\approx 6.3$ million samples. We see that all of these channels are highly correlated with each other; i.e. there is not much information to be gained by considering each of these channels separately. Thus, we simply use the average of these 4 channels as our input to the nowcasting model.

\begin{table}[h!]
\centering
\setlength{\tabcolsep}{3pt}
\renewcommand{\arraystretch}{1.3}
\begin{tabular}{|c!{\vrule width 1pt}c!{\vrule width 1pt}c!{\vrule width 1pt}c!{\vrule width 1pt}c!{\vrule width 1pt}c|}
\toprule[1pt]
Channel & \textbf{IR\_097} & \textbf{IR\_108} & \textbf{IR\_120} & \textbf{IR\_134} \\
\midrule[1pt]
\textbf{IR\_097} & 1.000 & 0.915 & 0.908 & 0.937 \\
\textbf{IR\_108} & 0.915 & 1.000 & 0.997 & 0.983 \\
\textbf{IR\_120} & 0.908 & 0.997 & 1.000 & 0.990 \\
\textbf{IR\_134} & 0.937 & 0.983 & 0.990 & 1.000 \\
\bottomrule[1pt]
\end{tabular}
\vspace{10pt}
\caption{Channel Correlation Matrix}
\label{tab:correlation_matrix}
\end{table}

Another important feature of our workflow, is that we focus only on cloudy regions for training the model as cloud-free regions are clearly not relevant for rainfall estimation. We use the Otsu thresholding method \cite{otsu1975threshold} to segment cloudy regions as the foreground and set the radiance values of all cloud-free regions to the maximum pixel value present in the image. This utilizes the fact that cloudy regions have depressed radiance values in the Infrared channels. The two preprocessing steps of a) channel selection and averaging and b) focus on cloudy regions through foreground/background segmentation allow us to use a cGAN with relatively lower model size to train the model to implement the translation from a radiance image to a rainfall image. 

Normalization was applied to ensure consistent scaling across datasets. The Input data was normalized to the range $[-1, 1]$ by dividing pixel values by 150 and subtracting 1. Similarly, OPERA rainfall rates were scaled by dividing values by 5 and subtracting 1. This standardization facilitated better convergence during training. To address missing values, NaN and infinite values in the Input dataset were replaced with the maximum observed value, which indicates the absence of cloud cover in the region. For the OPERA dataset, missing values were replaced with zero, representing no rainfall activity. To ensure compatibility with the neural network’s convolutional layers, reflection padding was applied to expand the $252 \times 252$ resolution data to $256 \times 256$. This step preserved edge information during convolution operations.

Sequence preparation was performed to capture temporal dynamics in the data. Each input sequence consisted of four consecutive SEVIRI frames representing one hour of satellite observations. The corresponding target sequence comprised sixteen OPERA rainfall frames, representing the subsequent four-hour rainfall prediction horizon.

This study utilizes the dense Lucas-Kanade optical flow algorithm as part of the current nowcasting pipeline; the algorithm being used has been implemented by OpenCV and then wrapped by the pySTEPS library \cite{opencv_library, pulkkinen_pysteps_2019}. We use this to predict the next 16 frames of HRIT radiance data from the given 4 input frames. The algorithm returns a motion vector for each detected feature within a given time series. This is not ideal as we require a corresponding motion vector for each pixel of HRIT image at the time of nowcast. The sparse motion vector field is then interpolated into a dense field using RBF (radial basis function) interpolation. There are multiple feature detection algorithms that can be used in conjunction with the Lucas-Kanade algorithm, we decided to use the blob feature detection method. 

The above method was used to propagate the radiance data forward, generating sixteen future frames from the initial four input frames. These predicted frames served as the input for the rainfall prediction model. This two-step approach ensured a seamless integration of spatio-temporal dependencies into the prediction process. The code repository for the described process is available on \href{https://github.com/flame-cai/Weather4Cast24_NIPS}{\underline{GitHub}} at \texttt{\url{https://github.com/flame-cai/Weather4Cast24\_NIPS}}.

\subsection{Model Design}
\label{subsec:model-des}
\begin{figure}[h!]
    \centering
    \fbox{\includegraphics[width=0.9\textwidth]{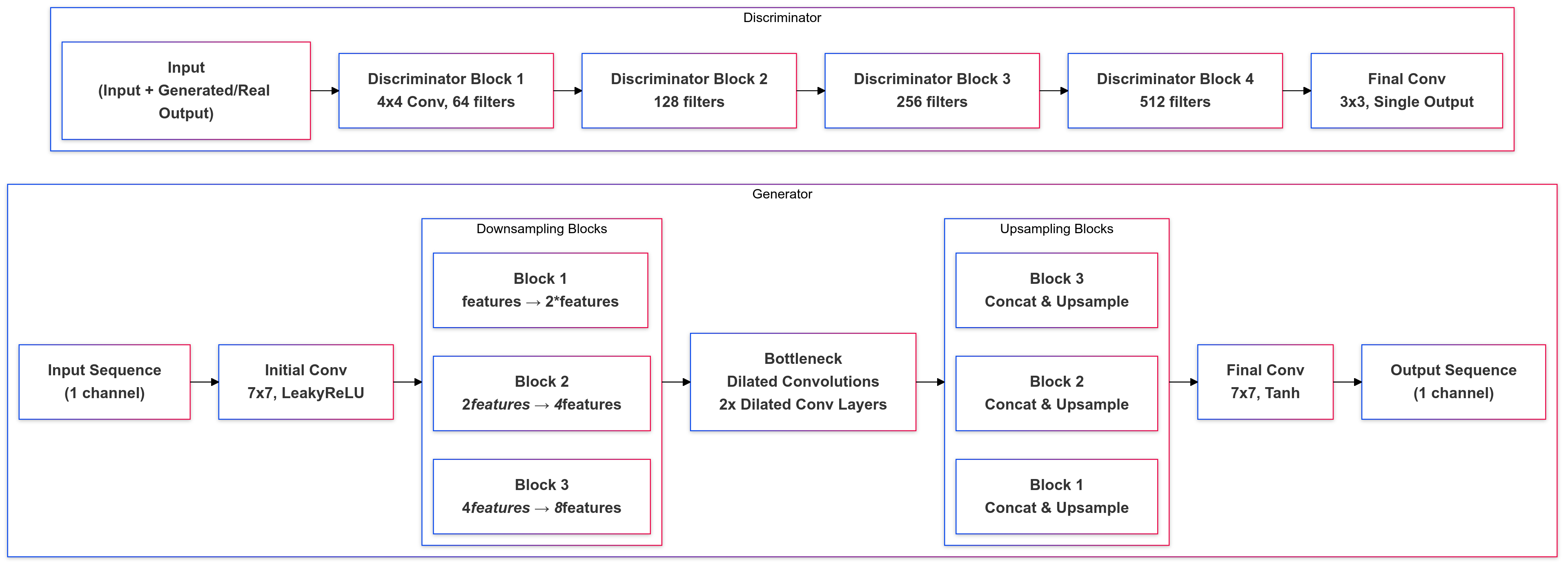}}
    \caption{Conditional GAN Architecture}
    \label{fig:block-diagram}
\end{figure}

The architecture is adapted from the Pix2Pix framework, enhanced with several modifications to improve performance for the rainfall prediction task. The pipeline consists of a generator and a discriminator, optimized using adversarial, pixel-wise, and perceptual loss functions \cite{isola_image-image_2016}. Figure~\ref{fig:block-diagram} depicts the block diagram of the different components of the model.

The \textbf{generator} employed a U-Net-inspired architecture with modifications tailored to the rainfall prediction task \cite{ayzel_rainnet_2020}. Down-sampling was performed using convolutional layers with instance normalization and LeakyReLU activations, compressing spatial dimensions while increasing feature depth. The bottleneck layer incorporated dilated convolutions to capture multi-scale spatial dependencies, which are critical for understanding rainfall patterns of varying intensities. Up-sampling layers employed transposed convolutions with skip connections, preserving spatial details and enhancing the quality of rainfall predictions.

The \textbf{discriminator} followed a PatchGAN architecture designed to classify input-output pairs as real or generated \cite{jia_patch-gen}. Sequential convolutional layers with filter sizes progressively increasing from 64 to 512 extracted meaningful features. Instance normalization and LeakyReLU activations facilitated efficient feature extraction. The final layer produced a single output value, representing the likelihood of authenticity for the predicted rainfall map.
 
\subsection{Training Procedure}
\label{subsec:train-proc}

The training process involved a single input frame as input and a single corresponding OPERA rainfall frame as output. The trained model was then applied to the input frames generated through optical flow to predict rainfall rates for the 16 future time steps. The generator was then trained to predict rainfall maps (OPERA) for each SEVIRI frame, effectively learning spatio-temporal relationships.

The training spanned 200 epochs with a batch size of 16. The Adam optimizer was used with $\beta_1 = 0.5$ and $\beta_2 = 0.999$, and a learning rate of $2 \times 10^{-4}$, which was scheduled cyclically to improve convergence. Loss functions included adversarial loss to ensure realistic predictions, pixel-wise loss (mean absolute error) for pixel-level accuracy, and perceptual loss using VGG-16 features to measure high-level structural similarity between predictions and ground truth. During training, the discriminator and generator were alternately updated to optimize their respective objectives.

The implementation was carried out in PyTorch and trained on an single NVIDIA RTX A4500 GPU. Custom data loaders were designed to handle sliding window sequences efficiently during training and validation. Optical flow predictions were seamlessly integrated with the rainfall prediction pipeline, ensuring smooth transitions between stages.

\section{Results}
\label{sec:results}
Due to the constraints of the competition format, this report only includes a simple visual analysis of the model results. Fig. 2 displays a sample predicted image to the left along with the corresponding reference image from OPERA to the right. The prediction for this example corresponds to a lead time of 4 hours. The model generally detects large rain bands - especially to the top left and bottom-centre of the image; though the shape and extent of the rain bands are not preserved exactly. The prediction misses fairly extensive rain regions in the bottom-right of the image. Further, the predictions severely underestimate the peak intensity of the rainfall relative to OPERA. Additionally, the model detects small patches of rain in the centre of the image while OPERA shows that these regions are rain-free. This is a predictable -- and perhaps inevitable -- consequence of using radiances related to cloud-top temperatures as a proxy for rain; we tend to misidentify rain-free regions with deep clouds as rainy areas.   

Despite these discrepancies, the cGAN model shows promise in identifying the general spatial distribution of rainfall, though several additional refinements are possible to improve its performance. We will discuss some of these in the next section.

\begin{figure}[h!]
    \centering
    \includegraphics[width=0.8\textwidth]{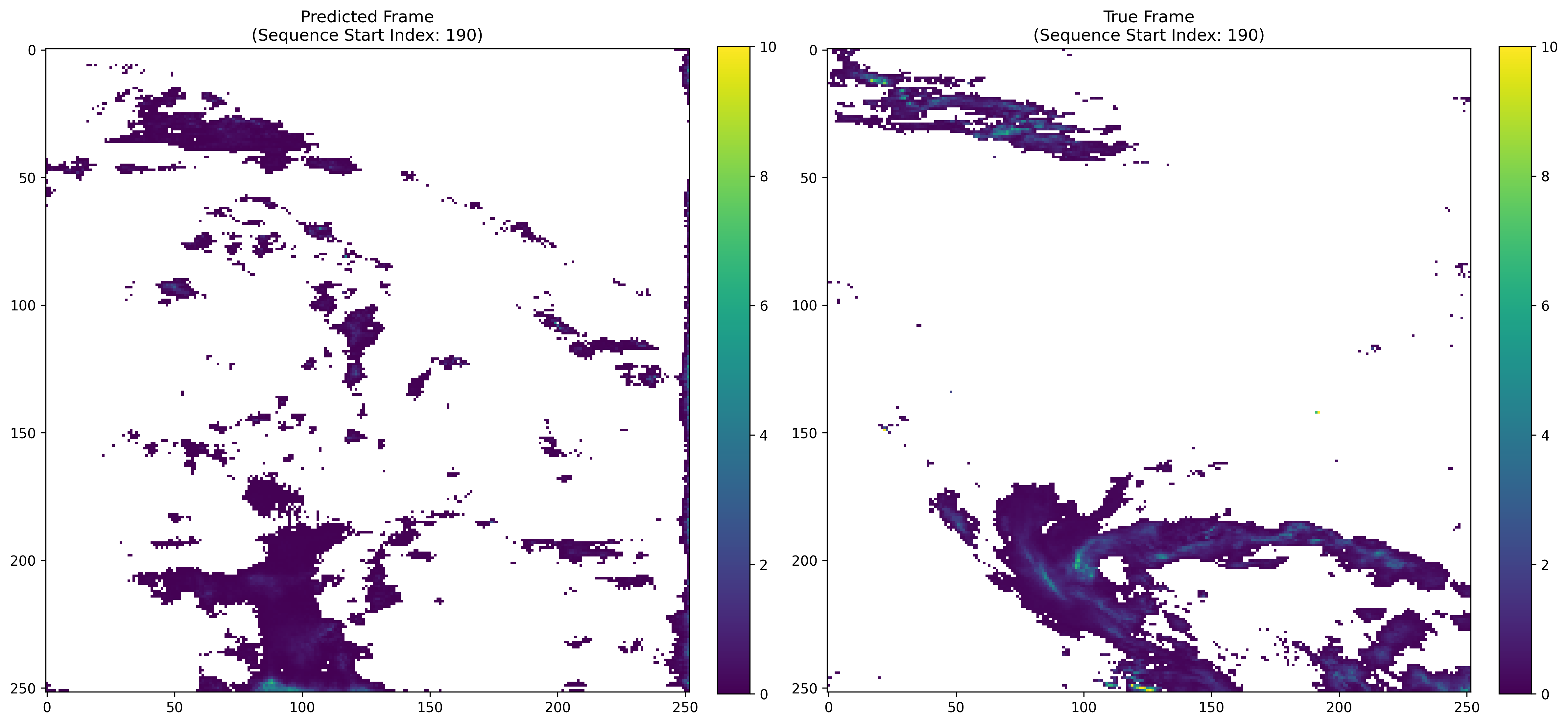}
    \caption{Sample predicted image after performing optical flow with a 2 hour lead time.}
    \label{fig:val-images}
\end{figure}

\begin{figure}[h!]
    \centering
    \begin{minipage}{0.32\textwidth}
        \centering
        \fbox{\includegraphics[width=\linewidth]{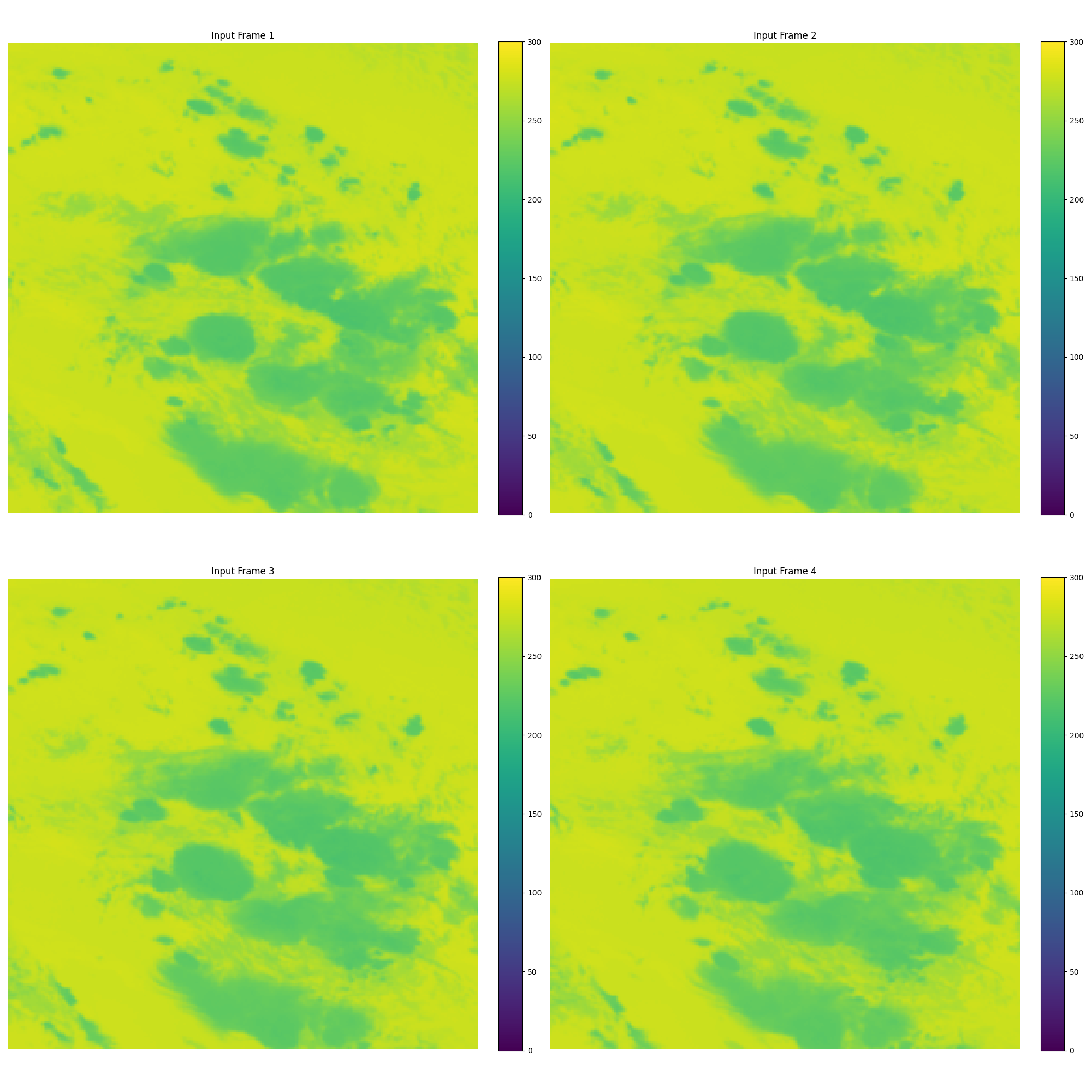}}
        \caption*{(a) Test input frames.}
    \end{minipage}
    \hfill
    \begin{minipage}{0.32\textwidth}
        \centering
        \fbox{\includegraphics[width=\linewidth]{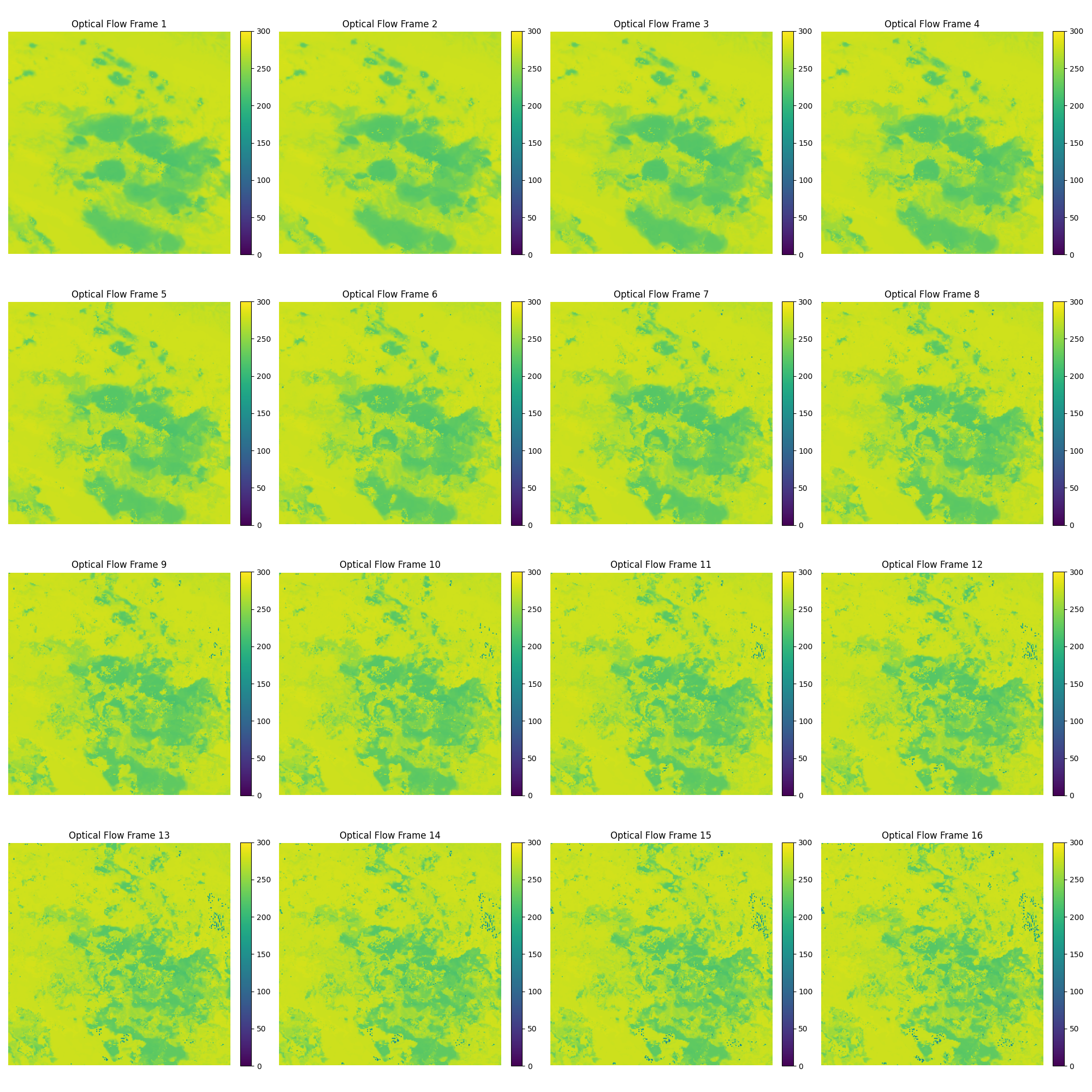}}
        \caption*{(b) Optical flow output.}
    \end{minipage}
    \hfill
    \begin{minipage}{0.32\textwidth}
        \centering
        \fbox{\includegraphics[width=\linewidth]{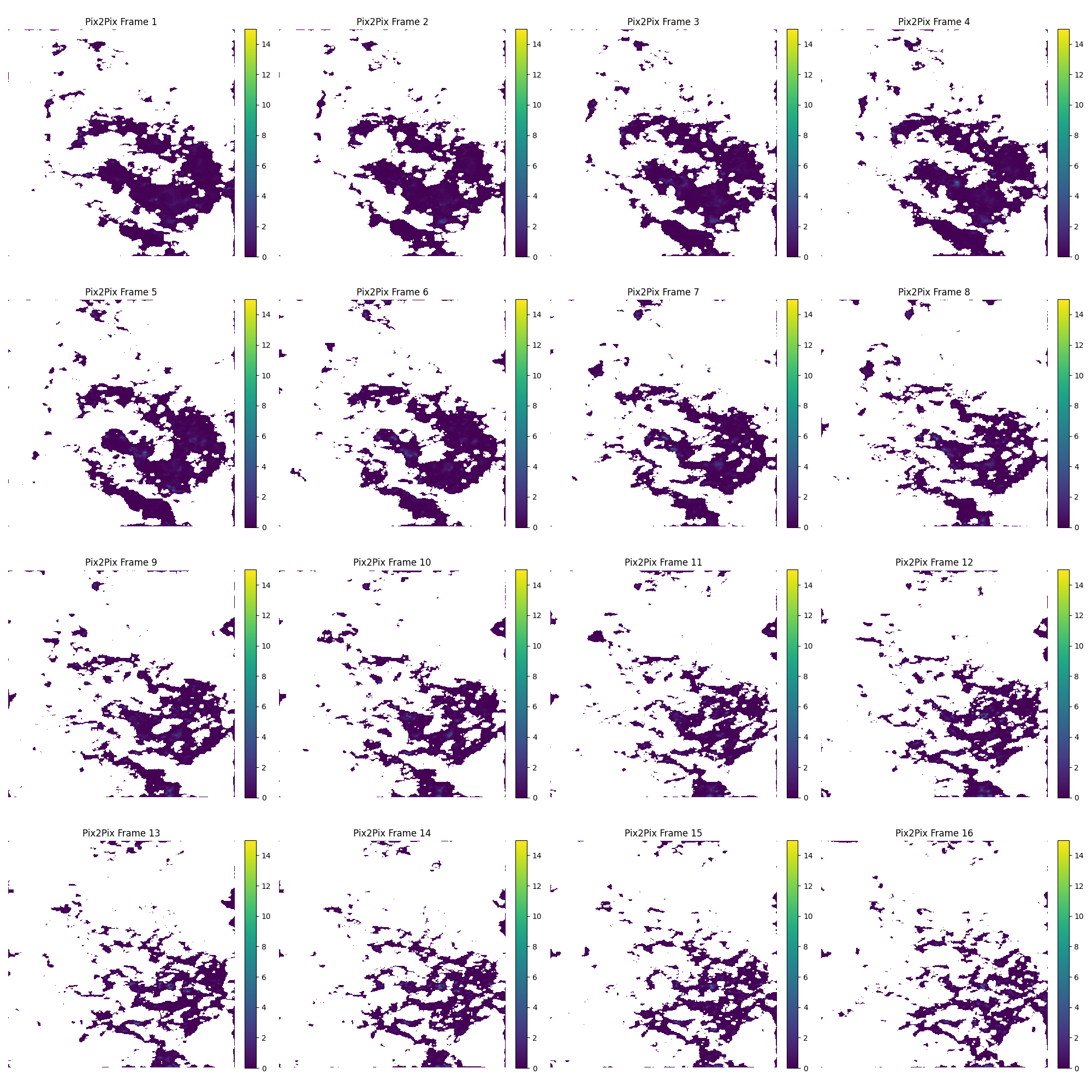}}
        \caption*{(c) cGAN output.}
    \end{minipage}
    \caption{Illustration of behavior of Optical flow algorithm}
    \label{fig:fig3}
\end{figure}

Next, we illustrate the behavior of the optical flow algorithm used in our submission (see fig.~\ref{fig:fig3}). The optical flow algorithm estimates the movement of the cloud system for 16 subsequent frames based on the average movement over the 4 input frames. We find that this method tends to cause increasing fragmentation in the estimated cloud systems, i.e large cloud systems in the input frames are seen to be ``breaking'' into smaller systems as the lead time for the optical flow prediction increases. We believe that this behavior -- which results in downstream fragmentation of the rain bands estimated by the cGAN model -- is an artifact of the averaging required to reconstruct the images from the dense optical flow vectors calculated in each frame.

The qualitative results are supported by the Continuous Ranked Probablity Score (CRPS) \cite{zamo_estimation_2018}, which was used to rank submissions in the Weather4Cast 2024 competition (See table~\ref{tab:crps-scores}) . The model described here was the only one to outperform the baseline model. While this is an encouraging result, we believe that this metric should be interpreted with caution, and we discuss some caveats in the next section.

\setlength{\arrayrulewidth}{0.5mm}
\setlength{\tabcolsep}{8pt}
\renewcommand{\arraystretch}{1.5}
\begin{table}[h!]
\centering
\begin{tabular}{|c|c|}
\hline
\textbf{Submissions} & \textbf{CRPS Score} \\ \hline
Proposed Model       & 7.34                \\ \hline
Best Baseline Model  & 10.84               \\ \hline
\end{tabular}
\captionsetup{skip=8pt}
\caption{Performance of the cGAN model in the competition leaderboard}
\label{tab:crps-scores}
\end{table}

\section{Discussion}
\label{sec:limit-future}

In this manuscript, we describe a deep learning workflow that consists of feature extraction, foreground/background segmentation, extrapolation using dense optical flow, and image translation using a cGAN model. While the procedure used for feature selection is admittedly ad-hoc in nature and can almost certainly be optimized to improve the results, it does allow a cGAN model of a fairly modest size to pick up relationships between the radiances and the rainfall values. The other innovation introduced in this manuscript is the use of foreground/background segmentation to mask out the variability in the cloud-free regions. This further allows the model to focus exclusively on the cloudy regions. While a basic validation exercise reveals decidedly mixed results, the metric chosen for the competition suggests that our model results in non-trivial improvements over the competition baseline. However, we wish to note here that submissions in the competition contain average rain rates over a 32 x 32 pixel area. Due to this, misalignments in the location and extent of the rain bands in the model estimates may not significantly affect the CRPS metric considered in the competition. 

Additionally, at this stage the translation from radiances to rainfall are considered independently at each time step. This is sub-optimal since the temporal variability of cloud-top temperatures is an important feature that can be used to evaluate the likelihood of rainfall occurence. We hope to develop more comprehensive deep learning models that can account for these temporal variations. Overall, the proposed approach demonstrates significant potential for use in nowcasting of rainfall, although significant improvements are required before it can be considered usable in an operational context.

\bibliographystyle{unsrt}  
\bibliography{references}

\end{document}